\documentclass[runningheads]{llncs}
\usepackage[T1]{fontenc}
\usepackage{graphicx}
\usepackage{subcaption}
\usepackage{booktabs}
\usepackage{siunitx}
\usepackage{amsmath}
\usepackage{amssymb}
\usepackage{tabularx} 
\usepackage{arydshln} 

\begin{document}
\title{Uniting contrastive and generative learning for event sequences models}
%
%
\author{Aleksandr Yugay \and
Alexey Zaytsev}
\authorrunning{A. Yugay and A. Zaytsev}
%
\institute{Skolkovo Institute of Science and Technology, Moscow, Russian Federation
\email{\{Aleksandr.Yugay, A.Zaytsev\}@skoltech.ru}\\
%
}
\maketitle              
\begin{abstract}
High-quality representation of transactional sequences is vital for modern banking applications, including risk management, churn prediction, and personalized customer offers. 
Different tasks require distinct representation properties: local tasks benefit from capturing the client's current state, while global tasks rely on general behavioral patterns. 
Previous research has demonstrated that various self-supervised approaches yield representations that better capture either global or local qualities. 

This study investigates the integration of two self-supervised learning techniques — instance-wise contrastive learning and a generative approach based on restoring masked events in latent space.
The combined approach creates representations that balance local and global transactional data characteristics. 
Experiments conducted on several public datasets, focusing on sequence classification and next-event type prediction, show that the integrated method achieves superior performance compared to individual approaches and demonstrates synergistic effects. 
These findings suggest that the proposed approach offers a robust framework for advancing event sequences representation learning in the financial sector.

\end{abstract}
\section{Introduction}

Transforming vast amounts of information into actionable insights enhances decision-making, optimizes operations, and improves customer experiences - this observation holds in the banking industry as well~\cite{ala2022deep,hassani2020deep}. 
Machine learning (ML) plays a critical role in this transformation, allowing financial institutions to handle complex tasks and extract valuable insights from their data~\cite{BANYMOHAMMED2024100215}.

Using models such as deep networks and hybrid models, banks accurately evaluate credit scores, predict financial distress, and detect risky transactions. 
This enhances their ability to assess risks and make informed lending decisions for a wide range of problems~\cite{BUENO2024100230}.




Recent developments in ML mostly concern the adoption of neural networks for financial data, including financial transactions modality~\cite{fursov2021adversarial}.
One of the reasons for their wide adoption is representation learning capabilities.
Data embedding is the core aspect of representation learning methods. It transforms raw data points or sequences into low-dimensional, fixed-length vectors. 
These vectors are designed to capture the underlying patterns or “nature” of the data, making them valuable for a wide range of downstream tasks~\cite{bazarova2024universal}.

Pre-trained embeddings from different domains are frequently employed for various purposes~\cite{devlin2018bert,he2022masked}.
They can be used as informative, ready-to-use input features for ML or deep learning models, reducing the need for extensive domain knowledge or significant feature engineering. 
Alternatively, these embeddings can serve as foundational building blocks when dealing with composite multi-modal data, where data from various sources or types are combined.
The most powerful methods here consider the self-supervised learning paradigm, where we train the encoder using unlabeled data~\cite{he2022masked}.

This idea is also adopted for financial transactions data~\cite{babaev2022coles,bazarova2024universal}.
For example, \cite{babaev2022coles} significantly advanced the field by introducing the CoLES approach based on the contrast between subsequences of different event sequences.
However, existing approaches suffer from the non-universality of embeddings: a model either focuses on a current moment to produce embeddings or provides a global picture, blurring important local details~\cite{bazarova2024universal}.
Part of the problem comes from the unavailability of a good local-scale model, as the direct adaptation of, e.g., Contrastive Predictive Coding~\cite{oord2018representation} for event sequences shows inferior results~\cite{babaev2022coles}.
The complexity here comes from the diverse information available in transactions. A single transaction event can include information about location, merchant category code, timestamp, amount, currency, etc.
Straightforward processing here works only if limited information about events available~\cite{moskvoretskii2024mlemgenerativecontrastivelearning}.

Our solution approaches both these problems: the lack of a good local representation learning model and the absence of universal, both local and global, properties in a single model.
We introduce CMLM (Contrastive Masked Language Model), a local approach that can deal with any input information, and a combination of CMLM with CoLES by proposing a weighted loss function. 
This approach concludes with the following contributions:
\begin{itemize}
    \item We propose CMLM based on the masking part of a sequence that can work with a broad range of available input information for each event.
    \item On top of it, we design a composition of CMLM and CoLES approaches by combining them in a single loss function. This loss is able to capture both local and global properties in a single embedding - and avoid the collapse of representations.
    \item The developed approach results in an encoder that produces representations with good local and global properties. This effect is evident in multiple financial transactions event sequence datasets.
\end{itemize}

\section{Related work}

Self-supervised learning (SSL) is a machine learning paradigm where models are trained using the data itself as labels, making predictions about one part of the data using other parts as a form of supervision~\cite{gui2024survey}. 
This approach is particularly valuable in leveraging large amounts of unlabeled data, which is often more readily available than labeled data. 
The motivation behind SSL arises from the data-hungry nature of deep learning algorithms, which require vast amounts of labeled data to avoid overfitting and biases. 
By using SSL, models can improve their generalization capabilities and handle out-of-distribution scenarios more effectively. There are two main types of self-supervised frameworks: contrastive and generative~\cite{liu2021self}. 

Contrastive methods in self-supervised learning involve learning representations by distinguishing between different instances in the data, often through data augmentation techniques that create pairs of similar and dissimilar examples. 
These approaches are particularly popular in the computer vision (CV) domain, with prominent examples including SimCLR~\cite{chen2020simple,chen2020big}, MoCo~\cite{chen2020improved,chen2021empirical,he2020momentum}, and SwAV~\cite{caron2021unsupervised}. 
Unlike traditional contrastive methods that rely on negative samples, approaches such as BYOL~\cite{grill2020bootstrap}, Barlow Twins~\cite{zbontar2021barlow}, and VICReg~\cite{bardes2022vicreg} eliminate the need for negative examples, focusing instead on maximizing similarity between different views of the same data instance.

Generative methods, on the other hand, aim to capture the underlying data distribution by generating new data instances or predicting missing parts.
In the NLP domain, models like BERT~\cite{devlin2018bert,lan2020albert,liu2019roberta} are trained to predict masked tokens using bidirectional context, enhancing their understanding of language structure. 
Similarly, the GPT~\cite{radford2018improving} models focus on autoregressive tasks, predicting the next token in a sequence. 
In the CV domain, analogous methods such as BEiT~\cite{bao2021beit} and MAE~\cite{he2022masked} have emerged, employing mask-based strategies to learn visual representations by predicting occluded parts of an image.

Each framework has its strengths and weaknesses. 
Previous research has shown that generative approaches excel in capturing local patterns, whereas contrastive methods provide higher-level abstractions beneficial for a comprehensive understanding of the object~\cite{qi2023contrast}. 
These frameworks have been adapted to transactional sequences~\cite{babaev2022coles,bazarova2024universal,moskvoretskii2024mlemgenerativecontrastivelearning}, where global tasks require a holistic understanding of the entire sequence, and local tasks focus on timestamp-specific details. 
The findings revealed that contrastive methods performed better on global tasks, while generative approaches excelled in addressing local tasks, consistent with previous research in the CV domain. 

It is tempting to develop a model that leverages the strengths of both approaches. 
However, prior research indicates that straightforward multitask training - i.e., simply combining the two losses - is ineffective. 
As a result, existing work either struggles to integrate the two approaches effectively or resorts to complex methods involving multiple models to achieve this integration~\cite{moskvoretskii2024mlemgenerativecontrastivelearning,qi2023contrast}. 
Thus, while both contrastive and generative methods offer distinct advantages, combining them remains a challenging task.


\section{Methods}

\subsection{Problem statement}
\paragraph{Self-supervised Learning (SSL).} Let \( X = \{ x_1, ..., x_N \} \) denote an unlabeled set of event sequences. 
Each sequence \( x_i = \{e_{ij}\}_{j=1}^{T_i} \) represents a temporal sequence of events, where \( e_{ij} \) denotes the \( j^{th} \) event in the \( i^{th} \) sequence, and \( T_i \) represents the length of the \( i^{th} \) sequence. 
Each event \( e_{ij} \) is characterized by multiple categorical or continuous features. 
It can include event type as well --- and all events are sorted in the sequence according to the occurrence time.

The objective, given \( X \), is to train a neural network \( f_{\theta} : \mathcal{X} \rightarrow \mathcal{Z} \), parameterized by \( \theta \), to map sequences to representations in a latent space. These representations should effectively capture the temporal dynamics and underlying structure inherent in the sequences.
Subsequently, representations acquired through \( f_{\theta} \) can be applied across diverse downstream tasks, including classification, regression, and clustering. 

This research focuses solely on evaluating the quality of representations for two classification tasks.
 The evaluation of learned representations involves assessing their performance on two distinct types of downstream tasks~\cite{bazarova2024universal}. 
This approach allows for the examination of different aspects of learned embeddings: global, which characterizes the sequence as a whole, and local, which captures the current state of individual instances. 
We can also fine-tune \( f_{\theta} \) for specific tasks or integrate it into larger models to enable multimodal capabilities.

\subsection{Model architecture}

A neural network model \(f_\theta\) consists of two main components: an event encoder \(g_{\theta_g}\) and a sequence encoder \(s_{\theta_s}\). The event encoder maps individual events into an embedding space.
In this paper, we focus on financial transactions data.
For such data modality, each event is characterized by two features: the Merchant Category Code (MCC), the amount and the event time. 
The MCC, being a categorical variable, is transformed into a $k$-dimensional embedding for each category. 
The amount, a continuous variable, is then concatenated with this MCC embedding to form the final embedding of the event. 
For other event sequences data, we see a similar pattern with a mix of categorical and continuous variables or a single characteristics of event presented.
So, our model and methods would benefit any event sequences data processing task.

\begin{figure}
  \centering
  \includegraphics[width=.7\linewidth]{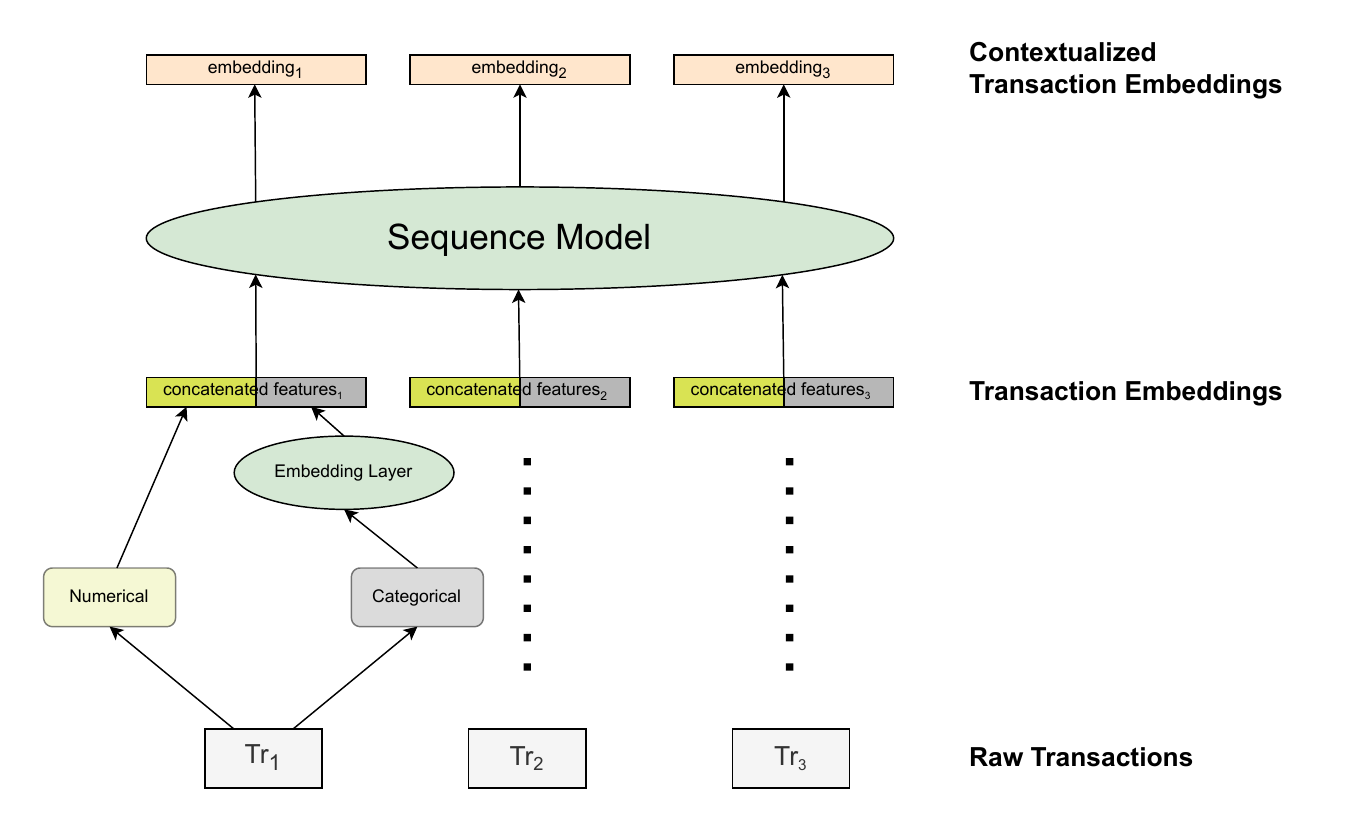}
  \caption{Diagram of a neural network model showing the transformation from raw transactions to contextualized transaction embeddings.}
  \label{fig:trans_model}
\end{figure}

The full process is illustrated in Figure~\ref{fig:trans_model} and follows similar methods described in \cite{10.1145/3292500.3330693,Udovichenko_2024}. 
Formally, for a sequence \(x = \{e_j\}_{j = 1}^T\), where \(e_j\) represents the \(j\)-th event, the event encoder \(g_{\theta_1}\) maps each event to its corresponding representation \(r_j\):
\[
\{e_j\}_{j = 1}^T \xrightarrow{g_{\theta_g}} \{r_j\}_{j = 1}^T.
\]
Subsequently, the sequence encoder \(s_{\theta_s}\) processes these event representations to generate contextualized representations or hidden states \(h_j\):
\[
\{r_j\}_{j=1}^T \xrightarrow{s_{\theta_s}} \{h_j\}_{j=1}^T.
\]

In this study, we used a unidirectional single-layer LSTM~\cite{hochreiter1997long} as the sequence encoder. The performance of this model is similar to transformers for the event sequences data modality, so we opt for a simpler model usage. 
To represent the entire sequence, we applied an aggregation function to the hidden states. Specifically, we used the hidden state of the last event as the sequence embedding, as it encapsulates all prior information~\cite{bazarova2024universal}.

\subsection{Contrastive learning with CoLES}

CoLES~\cite{babaev2022coles} is a contrastive self-supervised approach.
It tries to make positive pairs of subsequences close to each other in the embedding space, while for negative pairs it aims for the opposite effect.
There, positive pairs are subsequences of event sequences, while negative pairs are slices from different sequences. 
Moreover, it uses hard-negative mining~\cite{robinson2021contrastive} to use the most challenging negative pairs from a batch --- improving the learning process. 

The loss function for CoLES is defined as follows:
\[
\mathcal{L}^{\text{CoLES}} =\!\! \sum_{x^+ \in P(x)} d^2\left(f_{\theta}(x), f_{\theta}(x^+)\right) 
+ \!\! \sum_{x^- \in N(x)} \max\left\{0, \rho - d\left(f_{\theta}(x), f_{\theta}(x^-)\right)\right\}^2.
\]
Here, $d$ is the cosine distance function, $P(x)$ and $N(x)$ represent the sets of positive and negative samples for sample $x$ respectively from a batch, and $\rho > 0$ is a hyperparameter denotes the distance margin for negative pairs. 

By making embeddings of different parts of a single sequence close, CoLES tends to produce embeddings that describe sequences as a whole, representing \emph{global properties of an event sequence}.

\subsection{Our CMLM}

We refer to our approach as Contrastive Masked Language Modeling (CMLM).
It combines ideas from Contrastive Predictive Coding~\cite{oord2018representation} (CPC) and Masked Language Modeling~\cite{devlin2018bert} (MLM) to learn representations of sequential data. 

MLM is a self-supervised learning technique introduced in the NLP domain. 
A portion of tokens in a sequence is randomly masked with a special token, and the model's objective is to predict the original tokens based on the context provided by the remaining tokens. 
This prediction task is a classification problem because the tokens are discrete entities that the model must identify from a finite set of possible tokens, so cross-entropy loss is commonly used.

However, MLM cannot be directly applied to transactional sequences because transactions are not discrete tokens; they consist of various categorical and numerical features. 
To adapt MLM to predict transactions, a common approach in the literature involves training multiple prediction heads, each responsible for predicting a different feature of the transaction~\cite{bazarova2024universal,moskvoretskii2024mlemgenerativecontrastivelearning,Udovichenko_2024}. 
The loss function is then constructed as the sum of cross-entropy losses for categorical features and mean squared errors for numerical features, allowing the model to handle the different types of features appropriately. 

While the approach of using multiple prediction heads for transactional data offers a tailored method for handling diverse features, it presents several limitations. 
One major challenge is the difficulty in capturing the interdependencies between different features. 
Transactions often involve complex relationships among categorical and numerical features and treating each feature independently may lead to suboptimal modeling of these interactions. 
Additionally, the approach requires aggregating multiple loss functions — such as cross-entropy for categorical features and mean squared error for numerical features — which can complicate the training process and necessitate careful weighting strategies to avoid bias towards certain types of features. 

Instead of employing multiple prediction heads to model each transaction feature individually, we examine an alternative approach inspired by CPC, focusing on predicting transactions in the latent space.
By focusing on predicting the embedding of transactions rather than each transaction's individual features, the model aims to capture a compact representation that encompasses the essential information while abstracting away from specific details. 
For instance, predicting MCCs directly can be challenging due to subtle differences between similar codes. 
However, such transactions can be mapped close together in the embedding space, allowing the model to understand their shared context without explicitly predicting the MCC, thus simplifying the modeling process.



Integrating concepts from both MLM and CPC, the CMLM approach randomly masks some events in the input sequence with a special token and then forces the model to correctly predict the embeddings of these masked events. 
So, the loss function is defined as:
\begin{equation}
    \mathcal{L}^{\text{CMLM}} = -\sum\limits_{i}\log \dfrac{e^{\text{sim}(r_i, \hat{r}_i)}}{e^{\text{sim}(r_i, \hat{r}_i)} + \sum\limits_{j \in J}e^{\text{sim}(r_i, \hat{r}_j)}},
\end{equation}
where \(r_i\) is the embedding of a masked transaction, \(\hat{r}_i\) is the predicted embedding of the masked transaction, \(\text{sim}(\cdot, \cdot)\) denotes cosine similarity, and \(J\) is a set of randomly chosen indices serving as negative samples.

These works~\cite{oord2018representation,poole2019variational} show that minimizing this loss is equivalent to maximizing a lower bound on the mutual information between high-dimensional data and their representations. 
On the other hand, by construction, produced embeddings should describe \emph{local properties} of the model.

\subsection{Hybrid approach: CMLM + CoLES}

\begin{figure}[t!]
  \centering
  \includegraphics[width=.6\linewidth]{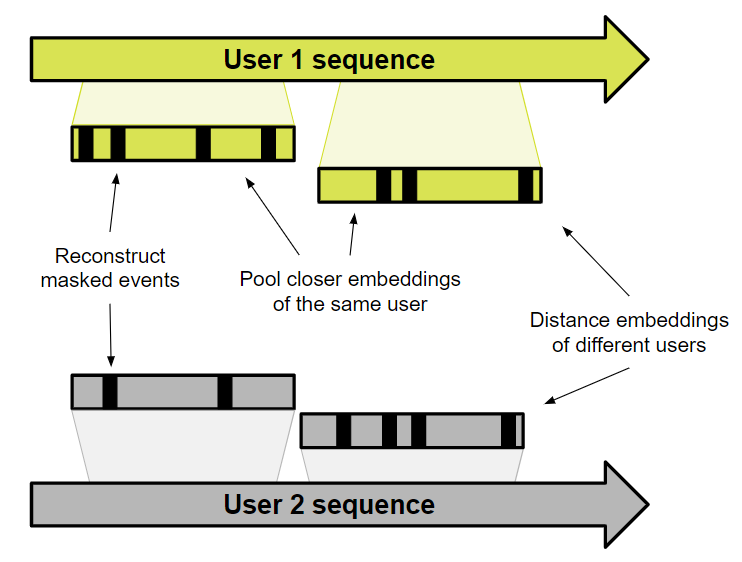}
  \captionof{figure}{Scheme of our CoLES+CMLM approach}
  \label{fig:cmlm_coles}
\end{figure}



The approaches described above belong to two different paradigms of self-supervised learning for sequential data: CoLES tends to encode global properties, while our CMLM focuses on local properties.

So, we propose a hybrid approach that combines CMLM's task of predicting masked events in latent space with CoLES's objective of distinguishing between different users in the simplest way possible. 
The loss function is a straightforward combination of two losses:
\begin{equation}
    \mathcal{L}^{\text{CMLM+CoLES}} = \mathcal{L}^{\text{CoLES}} + \lambda \mathcal{L}^{\text{CMLM}},
\end{equation}
where the factor \(\lambda\) balances the contributions of CoLES and CMLM to the overall loss.
The scheme of our approach is in Figure~\ref{fig:cmlm_coles}.

\section{Experiments}

\subsection{Datasets}

To ensure thorough evaluation, we tested the proposed methods on four public transactional datasets from various data science competitions, following existing benchmarks~\cite{babaev2022coles,bazarova2024universal}. 
Each dataset features transactions with details such as user ID, date, MCC, amount, and other relevant attributes. Descriptive statistics are available in Table~\ref{tab:dataset_stats}, with more details provided in the next paragraph.

\emph{Churn (Rosbank)} aims at predicting the probability of customer churn. The binary target labels are nearly balanced. The data spans from October 2016 to April 2018.
\emph{Gender (Sber)} involves the prediction of a client's gender based on their transactional activity. It covers the period from January 2022 to April 2023.
\emph{Age (Sber)} has the age group label. The multiclass target labels are evenly distributed across four age groups.
\emph{DataFusion (VTB)} has the objective to predict a bank customer churn by developing an algorithm that forecasts the likelihood of churn in the subsequent six months. The dates' interval is from October 2021 to March 2023.

\begin{table}[t!]
    \centering
    \caption{Dataset Statistics}
    \label{tab:dataset_stats}
    \begin{tabularx}{\textwidth}{l >{\raggedleft\arraybackslash}X >{\raggedleft\arraybackslash}X >{\raggedleft\arraybackslash}X >{\raggedleft\arraybackslash}X}        \toprule
        & \textbf{Churn} & \textbf{Gender} & \textbf{Age} & \textbf{DataFusion} \\
        \midrule
        Num. of Transactions & 490K & 2.9M & 26M & 8.7M \\
        Num. of Sequences & 5K & 7.4K & 30K & 64K \\
        Mean Seq. Length & \num{98.1} & \num{388.2} & \num{881.7} & \num{136.5} \\
        Std. Seq. Length & \num{78.1} & \num{309.4} & \num{124.8} & \num{148.9} \\
        Num. of Unique MCC & \num{344} & \num{184} & \num{202} & \num{323} \\
        \bottomrule
    \end{tabularx}
\end{table}

\begin{table}[t!]
    \centering
    \caption{Model hyperparameters}
    \label{tab:model_params}
    \begin{tabularx}{\textwidth}{l >{\raggedleft\arraybackslash}X >{\raggedleft\arraybackslash}X >{\raggedleft\arraybackslash}X >{\raggedleft\arraybackslash}X}
    \toprule
     & \textbf{Churn} & \textbf{Gender} & \textbf{Age} & \textbf{DataFusion} \\
    \midrule
    Embedding size & 24 & 24 & 24 & 24 \\
    Vocabulary size & 344 & 184 & 202 & 323 \\
    Hidden size & 512 & 128 & 512 & 64 \\
    Number of epochs & 50 & 100 & 50 & 50\\ 
    \bottomrule
    \end{tabularx}
\end{table}

To assess both global and local properties of the obtained embeddings, we conducted two distinct downstream tasks for each dataset: sequence classification, as for each dataset we have a corresponding target, and prediction of the next transaction's MCC. For sequence classification, we used CatBoost~\cite{10.5555/3327757.3327770} as the downstream model folllowing~\cite{babaev2022coles}, as typically results are superior to linear probing and fully connected neural networks. Meanwhile, for the next event type prediction, we trained a linear layer atop of a frozen encoder, following the approach described in~\cite{bazarova2024universal} --- again, here we adopt the best practice available in the literature. 
We evaluated the performance using the ROC-AUC metric, applying a weighted average for non-binary target variable.

All experiments were conducted using PyTorch~\cite{paszke2019pytorch} and executed on an NVIDIA GeForce RTX 3060 GPU. 
The datasets were split into training, validation, and test sets with an 80/10/10 ratio to provide a robust evaluation framework. 
Each experiment was run 5 times to account for variability and enhance the reliability of the results. 
A summary of the hyperparameters used is provided in Table~\ref{tab:model_params}.

\subsection{Results}

\begin{figure}[t!]
    \centering
    \includegraphics[width=\textwidth]{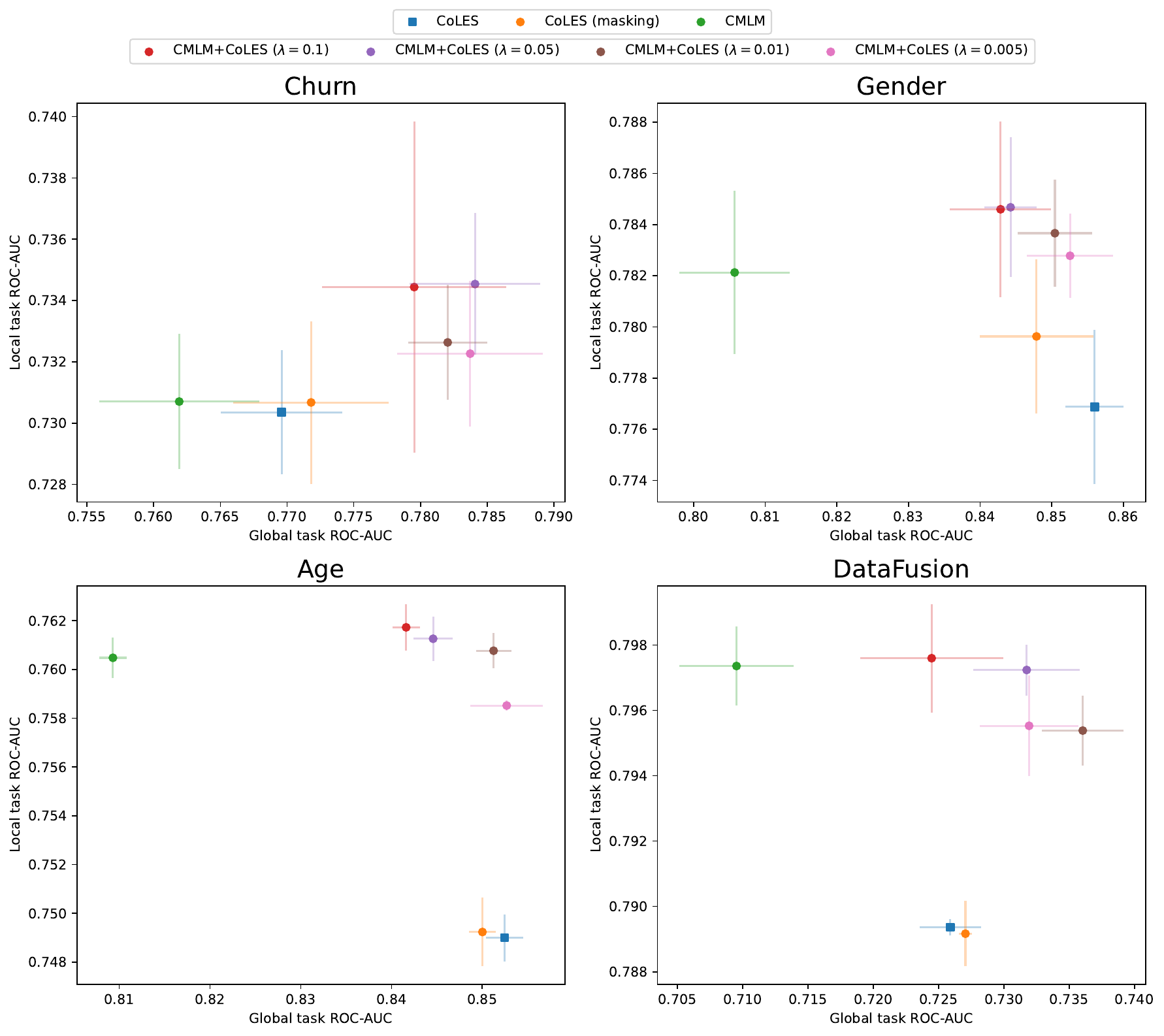}
    \caption{Quality metrics for global and local embedding tasks, showing \mbox{ROC-AUC} scores across four datasets.
    Each point represents the mean performance of a model, with error bars indicating standard deviation.}
    \label{fig:results}
\end{figure}

This study evaluated a hybrid representation learning approach against the baseline method, CoLES, across four datasets: Churn, Gender, Age, and DataFusion. 
The analysis considered both global and local performance, providing insights into how each method performs relative to CoLES in capturing both overall and timestamp-specific transaction sequence patterns.
The results are provided in Table~\ref{tab:results} and Figure~\ref{fig:results}. 
Additionally, Table~\ref{tab:results} compares our methods to two generative approaches - Masked Language Model (MLM) and Autoencoder (AE) - from~\cite{bazarova2024universal}.
For convenience, Table~\ref{tab:results_ranks} presents the mean performance ranks of the evaluated methods, highlighting our hybrid approach's strong results in both global and local tasks.


Compared to CoLES, CMLM improves in local performance but falls behind when considering global performance. This observation holds for all $4$ considered datasets. 
Thus, CMLM is effective in capturing the current state of individual transactions but struggles with generalizing across entire sequences. 
This aligns with previous research indicating that generative models excel at local tasks~\cite{bazarova2024universal}.

The proposed hybrid method CMLM+CoLES performed well across several datasets, capturing both global and local transaction patterns. 
In the Churn dataset, it outperformed CoLES in both global and local tasks. 
For the Gender dataset, it matched CMLM in local performance while maintaining comparable global performance to CoLES with $\lambda=0.005$. 
Similar patterns were observed in the Age and DataFusion datasets.
Additional experiments for CoLES with masking shows inferior results compared to this approach.
So, the combination of CMLM and CoLES effectively captures both global and local transaction patterns. 

The influence of the hyperparameter $\lambda$ is well-defined: increasing its value places greater emphasis on the CMLM loss, which enhances the model’s ability to capture local patterns but reduces its ability to capture global ones. 
This parameter thus creates a trade-off between the local and global properties of the resulting embeddings.

\begin{table}[t!]
    \centering
    \caption{ROC-AUC metrics for the validation results of global and local properties of embeddings. The results, averaged over five runs, are presented in the format mean $\pm$ standard deviation. The best values are \textbf{highlighted}, while the second-best values are \underline{underlined}.}
    \label{tab:results}
    \renewcommand{\arraystretch}{1.5} 
    \resizebox{\textwidth}{!}{ 
    \begin{tabular}{lcccccccc}
        \toprule
        & \multicolumn{2}{c}{\textbf{Churn}} & \multicolumn{2}{c}{\textbf{Gender}} & \multicolumn{2}{c}{\textbf{Age}} & \multicolumn{2}{c}{\textbf{DataFusion}} \\
        \cmidrule(lr){2-3} \cmidrule(lr){4-5} \cmidrule(lr){6-7} \cmidrule(lr){8-9}
        & Global & Local & Global & Local & Global & Local & Global & Local \\
        \midrule
        CoLES & 0.770{\tiny $\pm$0.007} & 0.730{\tiny $\pm$0.003} & \textbf{0.856{\tiny $\pm$0.005}} & 0.777{\tiny $\pm$0.004} & \underline{0.852{\tiny $\pm$0.002}} & 0.749{\tiny $\pm$0.001} & 0.726{\tiny $\pm$0.003} & 0.789{\tiny $\pm$0.000} \\
        AE & 0.756{\tiny $\pm$0.007} & 0.701{\tiny $\pm$0.004} & 0.676\tiny{$\pm$0.021} & 0.753\tiny{$\pm$0.002} & 0.782 \tiny{$\pm$0.010}& 0.722\tiny{$\pm$0.007}& 0.657\tiny{$\pm$0.013} &0.751\tiny{$\pm$0.009} \\
        MLM & 0.753\tiny{$\pm$0.014}
 & 0.723\tiny{$\pm$0.003} & 0.833\tiny{$\pm$0.004} & 0.764\tiny{$\pm$0.002} & 0.837\tiny{$\pm$0.006} & 0.714\tiny{$\pm$0.002} & 0.716\tiny{$\pm$0.011} & 0.766\tiny{$\pm$0.002}  \\
        \cdashline{1-9}
        CoLES (only masking) & 0.772{\tiny $\pm$0.007} & 0.731{\tiny $\pm$0.003} & 0.848{\tiny $\pm$0.009} & 0.780{\tiny $\pm$0.003} & 0.850{\tiny $\pm$0.001} & 0.749{\tiny $\pm$0.001} & 0.727{\tiny $\pm$0.001} & 0.789{\tiny $\pm$0.001} \\
        CMLM & 0.762{\tiny $\pm$0.010} & 0.731{\tiny $\pm$0.004} & 0.806{\tiny $\pm$0.009} & 0.782{\tiny $\pm$0.004} & 0.809{\tiny $\pm$0.002} & 0.760{\tiny $\pm$0.001} & 0.710{\tiny $\pm$0.005} & \underline{0.797{\tiny $\pm$0.001}} \\
        CMLM+CoLES {\scriptsize ($\lambda=0.1$)} & 0.780{\tiny $\pm$0.008} & \underline{0.734{\tiny $\pm$0.006}} & 0.843{\tiny $\pm$0.007} & \textbf{0.785{\tiny $\pm$0.004}} & 0.842{\tiny $\pm$0.002} & \textbf{0.762{\tiny $\pm$0.001}} & 0.724{\tiny $\pm$0.005} & \textbf{0.798{\tiny $\pm$0.001}} \\
        CMLM+CoLES {\scriptsize ($\lambda=0.05$)} & \textbf{0.784{\tiny $\pm$0.008}} & \textbf{0.735{\tiny $\pm$0.004}} & 0.844{\tiny $\pm$0.004} & \textbf{0.785{\tiny $\pm$0.003}} & 0.845{\tiny $\pm$0.002} & \underline{0.761{\tiny $\pm$0.001}} & 0.732{\tiny $\pm$0.005} & \underline{0.797{\tiny $\pm$0.001}} \\
        CMLM+CoLES {\scriptsize($\lambda=0.01$)} & \underline{0.782{\tiny $\pm$0.005}} & 0.733{\tiny $\pm$0.003} & 0.850{\tiny $\pm$0.005} & \underline{0.784{\tiny $\pm$0.002}} & 0.851{\tiny $\pm$0.002} & \underline{0.761{\tiny $\pm$0.001}} & \textbf{0.736{\tiny $\pm$0.003}} & 0.795{\tiny $\pm$0.001} \\
        CMLM+CoLES {\scriptsize ($\lambda=0.005$)} & \textbf{0.784{\tiny $\pm$0.009}} & 0.732{\tiny $\pm$0.004} & \underline{0.853{\tiny $\pm$0.008}} & 0.783{\tiny $\pm$0.002} & \textbf{0.853{\tiny $\pm$0.005}} & 0.759{\tiny $\pm$0.000} & \underline{0.734{\tiny $\pm$0.005}} & 0.795{\tiny $\pm$0.001} \\
        \bottomrule
    \end{tabular}
    }
\end{table}

\begin{table}[t!]
    \centering
    \caption{
    The mean model performance ranks for validation results of the global and local properties of embeddings.
    All ranks are computed with respect to the ROC-AUC metric and averaged over all datasets.
    The best values are \textbf{highlighted}, while the second-best values are \underline{underlined}.}
    \label{tab:results_ranks}
    \renewcommand{\arraystretch}{1.2}
    \begin{tabularx}{\textwidth}{l >{\raggedleft\arraybackslash}X >{\raggedleft\arraybackslash}X}
        \toprule
        & Global task & Local task  \\
        \midrule
        CoLES & \underline{2.000} & 3.750 \\
        AE  & 5.750 & 5.750 \\
        MLM & 4.500 & 5.250 \\
        \cdashline{1-3}
        CoLES (only masking) & 2.500  & 3.125 \\
        CMLM & 4.750 & \underline{1.875}  \\
        CMLM+CoLES {\scriptsize ($\lambda=0.01$)} & \textbf{1.500} & \textbf{1.250}   \\
        \bottomrule
    \end{tabularx}
\end{table}

\section{Conclusions}

We presented an approach that combines the strengths of contrastive and generative learning methods to improve representation learning for event sequences. 
The proposed CMLM approach solves a generative task — masked modeling — in the embedding space. 
This allows the model to focus on the core aspects of the data while ignoring less significant details.

Extensive experiments show, that CMLM outperforms CoLES in local pattern modeling. 
Furthermore, our hybrid approach CMLM+CoLES, which combines them through a simple multitask learning framework, achieved notable improvements in both local and global properties of learned representations. 
This simplicity contrasts with other complex hybrid methods proposed before. 
However, one limitation is the need to carefully select the hyperparameter \(\lambda\), which controls the balance between the generative and contrastive tasks.

In summary, our hybrid approach combining generative and contrastive learning methods significantly advances event sequence modeling.
Our universal method enhances both local and global representation learning while maintaining simplicity and flexibility. 
Future research could extend its application to various sequence types and domains.

\section{Acknowledgments}

The research was supported by the Russian Science Foundation grant 20-7110135. We also thank our lab team members for providing the code for the baselines.
 
\bibliographystyle{splncs04}
\bibliography{Bibliography}

\begin{thebibliography}{10}
\providecommand{\url}[1]{\texttt{#1}}
\providecommand{\urlprefix}{URL }
\providecommand{\doi}[1]{https://doi.org/#1}

\bibitem{ala2022deep}
Ala’raj, M., Abbod, M.F., Majdalawieh, M., Jum’a, L.: A deep learning model for behavioural credit scoring in banks. Neural Computing and Applications  \textbf{34}(8),  5839--5866 (2022)

\bibitem{babaev2022coles}
Babaev, D., et~al.: Coles: Contrastive learning for event sequences with self-supervision. In: SIGMOD. pp. 1190--1199 (2022)

\bibitem{10.1145/3292500.3330693}
Babaev, D., et~al.: {E.T.-RNN}: Applying deep learning to credit loan applications. In: ACM SIGKDD. p. 2183–2190. KDD '19, Association for Computing Machinery, New York, NY, USA (2019)

\bibitem{BANYMOHAMMED2024100215}
{Bany Mohammed}, A., Al-Okaily, M., Qasim, D., {Khalaf Al-Majali}, M.: Towards an understanding of business intelligence and analytics usage: Evidence from the banking industry. International Journal of Information Management Data Insights  \textbf{4}(1),  100215 (2024)

\bibitem{bao2021beit}
Bao, H., Dong, L., Piao, S., Wei, F.: {BEiT}: {BERT} pre-training of image transformers. In: ICLR (2022)

\bibitem{bardes2022vicreg}
Bardes, A., Ponce, J., Lecun, Y.: Vicreg: Variance-invariance-covariance regularization for self-supervised learning. In: ICLR (2022)

\bibitem{bazarova2024universal}
Bazarova, A., et~al.: Universal representations for financial transactional data: embracing local, global, and external contexts (2024)

\bibitem{BUENO2024100230}
Bueno, L.A., et~al.: Impacts of digitization on operational efficiency in the banking sector: Thematic analysis and research agenda proposal. International Journal of Information Management Data Insights  \textbf{4}(1),  100230 (2024)

\bibitem{caron2021unsupervised}
Caron, M., Misra, I., Mairal, J., Goyal, P., Bojanowski, P., Joulin, A.: Unsupervised learning of visual features by contrasting cluster assignments. NeurIPS  \textbf{33},  9912--9924 (2020)

\bibitem{chen2020simple}
Chen, T., Kornblith, S., Norouzi, M., Hinton, G.: A simple framework for contrastive learning of visual representations. In: ICML. pp. 1597--1607. PMLR (2020)

\bibitem{chen2020big}
Chen, T., Kornblith, S., Swersky, K., Norouzi, M., Hinton, G.E.: Big self-supervised models are strong semi-supervised learners. NeurIPS  \textbf{33},  22243--22255 (2020)

\bibitem{chen2020improved}
Chen, X., Fan, H., Girshick, R., He, K.: Improved baselines with momentum contrastive learning (2020)

\bibitem{chen2021empirical}
Chen, X., Xie, S., He, K.: An empirical study of training self-supervised vision transformers. In: CVPR. pp. 9640--9649 (2021)

\bibitem{devlin2018bert}
Devlin, J., Chang, M.W., Lee, K., Toutanova, K.: Bert: Pre-training of deep bidirectional transformers for language understanding. arXiv preprint arXiv:1810.04805  (2018)

\bibitem{fursov2021adversarial}
Fursov, I., et~al.: Adversarial attacks on deep models for financial transaction records. In: ACM SIGKDD. pp. 2868--2878 (2021)

\bibitem{grill2020bootstrap}
Grill, J.B., et~al.: Bootstrap your own latent-a new approach to self-supervised learning. NeurIPS  \textbf{33},  21271--21284 (2020)

\bibitem{gui2024survey}
Gui, J., et~al.: A survey on self-supervised learning: Algorithms, applications, and future trends. IEEE Transactions on Pattern Analysis and Machine Intelligence  (2024)

\bibitem{hassani2020deep}
Hassani, H., et~al.: Deep learning and implementations in banking. Annals of Data Science  \textbf{7},  433--446 (2020)

\bibitem{he2020momentum}
He, K., Fan, H., Wu, Y., Xie, S., Girshick, R.: Momentum contrast for unsupervised visual representation learning. In: CVPR. pp. 9729--9738 (2020)

\bibitem{he2022masked}
He, K., et~al.: Masked autoencoders are scalable vision learners. In: CVPR. pp. 16000--16009 (2022)

\bibitem{hochreiter1997long}
Hochreiter, S., Schmidhuber, J.: Long short-term memory. Neural computation  \textbf{9}(8),  1735--1780 (1997)

\bibitem{lan2020albert}
Lan, Z., et~al.: {ALBERT}: A lite {BERT} for self-supervised learning of language representations. In: ICLR (2019)

\bibitem{liu2021self}
Liu, X., et~al.: Self-supervised learning: Generative or contrastive. IEEE transactions on knowledge and data engineering  \textbf{35}(1),  857--876 (2021)

\bibitem{liu2019roberta}
Liu, Y., et~al.: Roberta: A robustly optimized bert pretraining approach (2019)

\bibitem{moskvoretskii2024mlemgenerativecontrastivelearning}
Moskvoretskii, V., et~al.: {MLEM}: Generative and contrastive learning as distinct modalities for event sequences (2024), \url{https://arxiv.org/abs/2401.15935}

\bibitem{oord2018representation}
Oord, A.v.d., Li, Y., Vinyals, O.: Representation learning with contrastive predictive coding. arXiv preprint arXiv:1807.03748  (2018)

\bibitem{paszke2019pytorch}
Paszke, A., et~al.: Pytorch: An imperative style, high-performance deep learning library. NeurIPS  \textbf{32} (2019)

\bibitem{poole2019variational}
Poole, B., et~al.: On variational bounds of mutual information. In: ICML. pp. 5171--5180. PMLR (2019)

\bibitem{10.5555/3327757.3327770}
Prokhorenkova, L., et~al.: Catboost: unbiased boosting with categorical features. In: NeurIPS. NIPS'18, Curran Associates Inc., Red Hook, NY, USA (2018)

\bibitem{qi2023contrast}
Qi, Z., et~al.: Contrast with reconstruct: Contrastive 3d representation learning guided by generative pretraining. In: ICML. pp. 28223--28243. PMLR (2023)

\bibitem{radford2018improving}
Radford, A., et~al.: Improving language understanding by generative pre-training  (2018)

\bibitem{robinson2021contrastive}
Robinson, J., Chuang, C.Y., Sra, S., Jegelka, S.: Contrastive learning with hard negative samples. In: ICLR (2021)

\bibitem{Udovichenko_2024}
Udovichenko, I., et~al.: {SeqNAS}: Neural architecture search for event sequence classification. IEEE Access  \textbf{12},  3898–3909 (2024)

\bibitem{zbontar2021barlow}
Zbontar, J., Jing, L., Misra, I., LeCun, Y., Deny, S.: Barlow twins: Self-supervised learning via redundancy reduction. In: ICML. pp. 12310--12320. PMLR (2021)

\end{thebibliography}

\end{document}